\documentclass[conference]{IEEEtran}
\IEEEoverridecommandlockouts
\usepackage{cite}
\usepackage{url}
\usepackage{amsmath,amssymb,amsfonts,amsthm}
\usepackage{tcolorbox}
\usepackage{algorithmic}
\usepackage{graphicx}
\usepackage{textcomp}
\usepackage{xcolor}
\usepackage{array}
\usepackage{graphicx}
\usepackage{colortbl}
\usepackage{array}
\usepackage{fancyhdr}
\begin{document}

\title{Enhancing Deep Learning Model Robustness through Metamorphic Re-Training\\
\thanks{\textsuperscript{*}This is a report submitted for the practicum course - \textit{``Advanced Testing of Deep Learning Models - Towards Robust AI''} which was conducted at the Chair of Software and Systems Engineering (Technical University of Munich). Supervisors: Simon Speth (TUM) and Vivek V. Vekariya (TUM).}}

\author{\IEEEauthorblockN{Youssef Sameh Mostafa - Matriculation Number: 03788260}
\IEEEauthorblockN{Karim Lotfy - Matriculation Number: 03788165}
\IEEEauthorblockN{Aziz Said Togru - Matriculation Number: 12532570 (LMU)}
\IEEEauthorblockA {\textit{Chair of Software \& Systems Engineering} \\
\textit{Technical University of Munich}\\
Munich, Germany}
}

\maketitle

\begin{abstract} This paper evaluates the use of metamorphic relations to enhance the robustness and real-world performance of machine learning models. We propose a Metamorphic Retraining Framework, which applies metamorphic relations to data and utilizes semi-supervised learning algorithms in an iterative and adaptive multi-cycle process. The framework integrates multiple semi-supervised retraining algorithms, including FixMatch, FlexMatch, MixMatch, and FullMatch, to automate the retraining, evaluation, and testing of models with specified configurations. To assess the effectiveness of this approach, we conducted experiments on CIFAR-10, CIFAR-100, and MNIST datasets using a variety of image processing models, both pretrained and non-pretrained. Our results demonstrate the potential of metamorphic retraining to significantly improve model robustness as we show in our results that each model witnessed an increase of an additional flat 17 percent on average in our robustness metric.
\end{abstract}

\section{Introduction}

Machine learning (ML) has made significant strides in various fields, including image recognition, natural language processing, and predictive analytics. Despite these advancements, ML models often struggle with robustness and generalization when deployed in real-world scenarios. This challenge is primarily due to the models' dependency on the quality and diversity of the training data, as well as their inability to handle unexpected data variations and anomalies. As a result, there is a growing need for methods that can enhance the robustness and adaptability of ML models, enabling them to perform reliably under diverse and unpredictable conditions.

One of the primary weak points in the current ML landscape is the over-reliance on supervised learning, which requires large amounts of labeled data. Labeling data is a time-consuming and costly process, and in many domains, such extensive labeling is impractical. Consequently, there is an increasing interest in semi-supervised learning, which leverages both labeled and unlabeled data to improve model performance. Semi-supervised learning has shown promise in reducing the dependency on labeled data, but challenges remain in optimizing these methods for robustness and adaptability.

Another critical issue is the lack of robustness in ML models. Many models perform well on test datasets but fail to generalize to real-world data, which can be noisy, incomplete, or contain variations not seen during training. This lack of robustness can lead to significant performance degradation when models are exposed to new environments or conditions. Therefore, developing techniques to improve model robustness is crucial for the reliable deployment of ML systems in real-world applications.

Metamorphic testing, a technique initially developed for software testing, can serve as a potential solution to these challenges through the use of metamorphic relations to generate new data samples and evaluate the model. Metamorphic relations (MRs) describe how the outputs of a program should change in response to specific transformations of the input this corresponds to changing the ground truth label. By applying MRs to the training data, we can generate new, synthetic data points that preserve essential properties of the original data while introducing variations. This approach can help expose the model to a broader range of scenarios, improving its robustness and generalization.

In this paper, we introduce the Metamorphic Retraining Framework, a novel approach that integrates metamorphic relations with semi-supervised learning algorithms. The framework aims to enhance the robustness and adaptability of ML models by iteratively applying MRs to the data and retraining the models in a multi-cycle fashion. We incorporate various state-of-the-art semi-supervised learning algorithms, such as FixMatch \cite{200107685}, FlexMatch \cite{211008263} , MixMatch \cite{90502249} , and FullMatch \cite{chen2023boostingsemisupervisedlearningexploiting}, to optimize the retraining process.

Our framework automates the entire retraining, evaluation, and testing pipeline, allowing for a systematic and efficient assessment of the impact of metamorphic retraining on model performance. To validate the effectiveness of our approach, we conducted extensive experiments on well-known image datasets, including CIFAR-10 \cite{krizhevsky2009learning}, CIFAR-100 \cite{krizhevsky2009learning}, and MNIST \cite{deng2012mnist}. We evaluated a wide variety of image processing models, both pretrained and non-pretrained, to ensure a comprehensive analysis.

The relevance of our project lies in its potential to address some of the most pressing issues in the field of ML. By reducing the dependency on labeled data and enhancing model robustness, our framework can contribute to developing more reliable and adaptable ML systems. The ability to perform well under diverse and unpredictable conditions is essential for the widespread adoption of ML technologies in critical applications, ranging from autonomous driving to healthcare and finance.

In summary, this paper presents a systematic approach to improving ML model robustness and performance through metamorphic retraining. By leveraging metamorphic relations and semi-supervised learning, we aim to provide a scalable solution to some of the fundamental challenges in the field, paving the way for more resilient and versatile ML applications.

\section{Motivation}
The motivation for this project arises from the critical challenges faced by deep learning (DL) models, such as overfitting, poor generalization, and sensitivity to input variations. These issues often undermine the reliability and robustness of DL models in real-world applications, necessitating innovative approaches to improve their performance \cite{8970483}. To address these challenges, we aim to achieve the following research goals and learning outcomes:

\subsection{Framework Development}
Our goal is to create a comprehensive and flexible framework for DL model retraining that uses metamorphic relation definitions from the metamorphic testing GeMTest framework and semi-supervised learning techniques in a robustness pipeline to ensure adaptability to different models and datasets.

\subsection{Impact of Metamorphic Testing}
We will evaluate how the integration of metamorphic tests affects the overall performance and robustness of the model compared to baseline techniques. This includes measuring the effectiveness of metamorphic tests in identifying model weaknesses and assessing the subsequent performance improvements after retraining with augmented data, focusing on metrics such as accuracy, precision, recall, and robustness scores.

\subsection{Algorithm Comparison}
We want to evaluate which of the most commonly used semi-supervised algorithms perform best when retraining with metamorphic test results to determine which algorithm is most effective for enhancing model robustness. Our research will compare different semi-supervised learning algorithms, such as FixMatch \cite{200107685}, FlexMatch \cite{211008263} , MixMatch \cite{90502249} and, FullMatch \cite{chen2023boostingsemisupervisedlearningexploiting}.

\subsection{Training Data Selection}
One of our key questions is whether to retrain models using only failed test cases, only passed test cases, or a combination of both. We aim to investigate which strategy yields the best improvements in model robustness and performance.

\subsection{Influence of Pretrained Models}
We aim to understand how the robustness of a model is affected by prior biases introduced during pretraining. This includes investigating whether pretrained models are naturally more robust or if their robustness can be further improved through metamorphic retraining.

\subsection{Model and Dataset Analysis}
Identifying which models and datasets benefit the most from our proposed framework is crucial. We will analyze the results to determine any correlations between model architectures, dataset characteristics, and the improvements observed.
\newline
\newline
By addressing these challenges, our project seeks to provide valuable insights and practical solutions for enhancing the robustness and generalization capabilities of DL models. This will ultimately lead to more reliable performance in diverse real-world scenarios.

\section{Background}

\begin{figure*}[t]
    \centering
    \includegraphics[width=\linewidth]{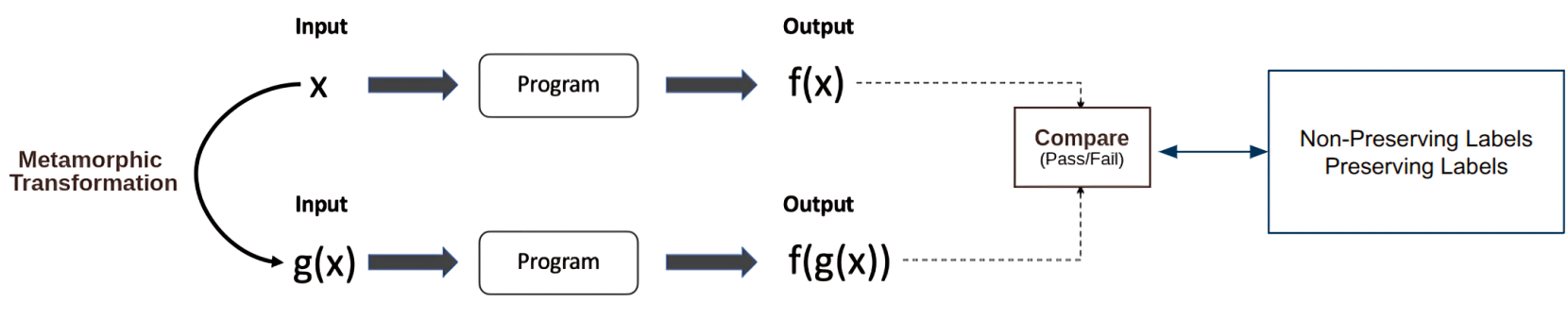}
    \caption{Metamorphic Testing Process \cite{metamorphic_testing}}
    \label{fig:metamorphic_testing}
\end{figure*}

\subsection{Adversarial Attacks and Neural Network Robustness}

Adversarial attacks present a critical challenge to neural network security by introducing small, often imperceptible perturbations to input data that mislead models into making incorrect predictions. These attacks exploit vulnerabilities in the decision boundaries of neural networks and can be executed under different knowledge settings: white-box attacks, where the attacker knows the model's details, and black-box attacks, where the attacker has no internal knowledge of the model \cite{madry2019deeplearningmodelsresistant}. Although our focus is not on generating adversarial attacks, understanding these concepts is essential because they highlight the need for robust neural networks that can withstand various input perturbations.

To combat these threats, enhancing network robustness is crucial. Robustness refers to a model's ability to maintain accuracy despite adversarial perturbations. Techniques like adversarial training, where models learn from adversarially perturbed examples, and defensive distillation, which trains a model on soft labels from a teacher model to smooth decision boundaries, are common methods to increase robustness. However, these techniques can sometimes reduce performance on unperturbed data and may not generalize across attack types \cite{9348948}.

Regularization techniques such as weight decay, dropout, and batch normalization help improve robustness by preventing overfitting and promoting smoother decision boundaries. Yet, achieving robustness involves navigating trade-offs between model accuracy and resistance to attacks, as well as addressing the transferability of adversarial examples between different models. Recent research focuses on developing methods that provide certified robustness, offering guarantees that models can withstand perturbations within specific bounds. Techniques like randomized smoothing and interval bound propagation are examples of this approach, aiming to enhance the resilience of neural networks under adversarial conditions \cite{190202918} \cite{181012715}.

Our work aims to enhance robustness against a different type of input variation—those captured by metamorphic relations. Metamorphic testing evaluates the robustness and reliability of neural networks by systematically transforming inputs and assessing model responses. While adversarial attacks and metamorphic testing both strive to improve model reliability under diverse conditions, they address different types of robustness. Metamorphic relations focus on benign input variations, whereas adversarial attacks target malicious perturbations.

Therefore, while the concepts and techniques for enhancing robustness in the face of adversarial attacks provide a valuable context, our primary focus remains on improving model reliability through metamorphic testing. This approach addresses robustness from a different angle, complementing existing methods by ensuring models can handle a variety of input transformations effectively.

\subsection{Metamorphic Testing and Relations}

Metamorphic testing is a powerful method for assessing the robustness of neural networks, particularly valuable in situations where traditional test oracles do not exist. This approach utilizes MRs, which define expected changes in output as a result of specific modifications to the input. This is crucial in scenarios where generating sufficient labeled data is problematic.

Metamorphic testing begins by identifying suitable MRs for the application at hand. These relations guide the creation of new test cases by altering existing inputs, thus creating scenarios to further evaluate the model. For example, transformations might involve rotating an image or changing the dataset's input parameters, with the expectation that the output adjusts in a predictable manner. By applying these transformations and observing the model’s responses, inconsistencies or weaknesses in the model can be detected.

This technique is particularly effective at uncovering errors that might escape traditional testing due to its ability to generate extensive test cases from minimal initial data. It is therefore highly beneficial in fields where data labeling is expensive or impractical \cite{210404718}.

The diagram in Figure \ref{fig:metamorphic_testing} visually summarizes the metamorphic testing process. An initial input \( x \) undergoes a metamorphic transformation to become \( g(x) \), and both versions are processed by the program. The outputs, \( f(x) \) and \( f(g(x)) \), are then compared to determine if the program behaves as expected under the defined MRs.

Metamorphic relations can be categorized into two types: label-preserving and non-label-preserving. 

\textbf{Label-Preserving MRs:} These maintain the output label despite changes to the input. For instance, in image recognition, rotating an image of a cat should still result in the label "cat." This tests the model's invariance to rotational transformations.

\textbf{Non-Label-Preserving MRs:} These involve changes that should predictably alter the output label. For example, changing a command from "turn left" to "turn right" in a navigation system should correctly change the direction in the output.

Overall, metamorphic testing not only enhances the reliability of neural networks by identifying and correcting vulnerabilities but also ensures that models perform robustly in varied real-world scenarios. It thus serves as a crucial tool in developing advanced neural network models that are both robust and generalizable.

\subsection{Metamorphic Testing Framework}

The GeMTest metamorphic testing framework is pivotal for validating DL models in scenarios where conventional testing methods fall short, particularly when outputs cannot be directly verified.

\begin{figure}[ht]
    \centering
    \includegraphics[width=\linewidth]{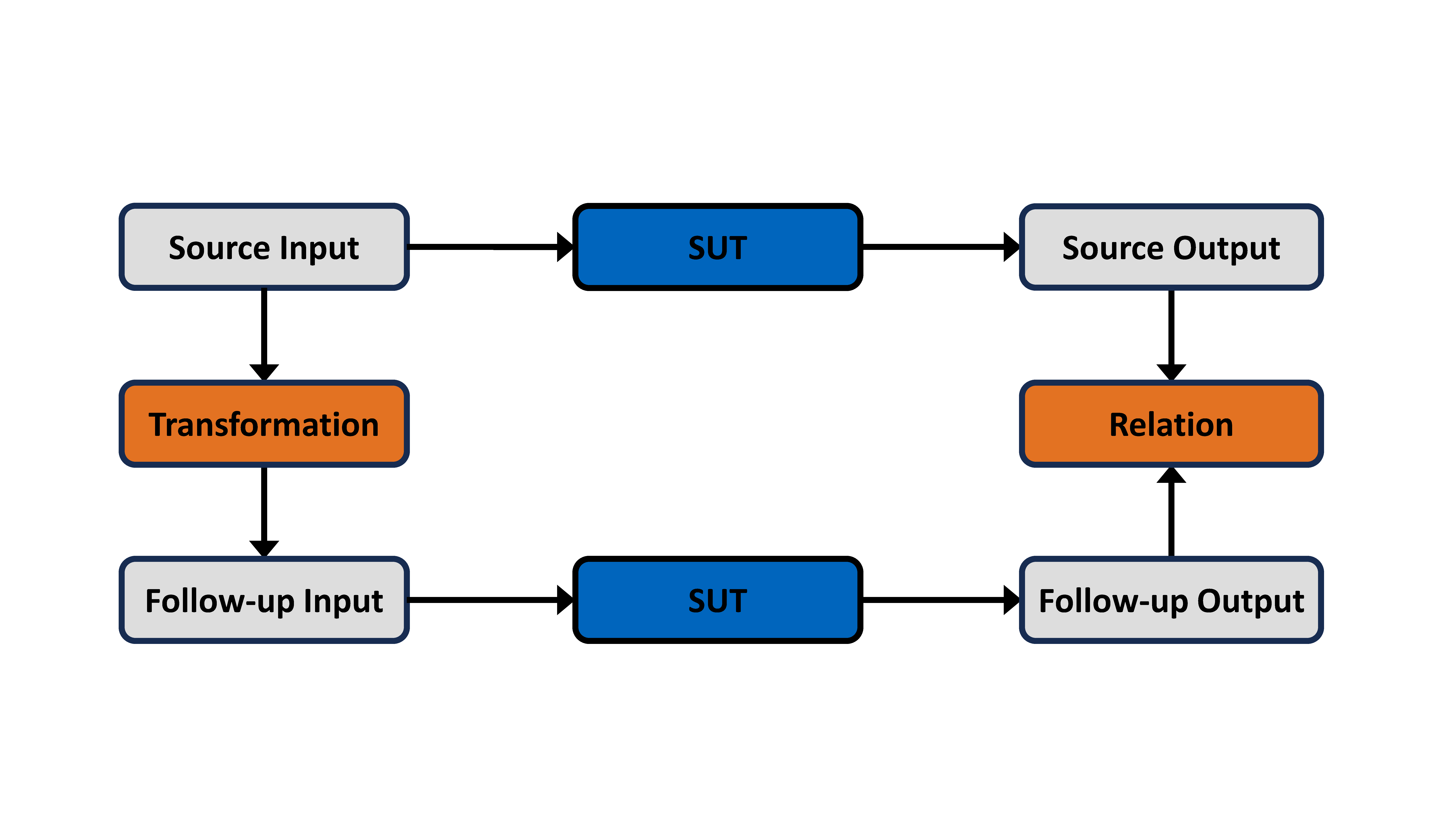}
    \caption{Workflow of the Metamorphic Testing Framework}
    \label{fig:metamorphic_framework}
\end{figure}

GeMTest employs metamorphic relations to generate test cases by transforming source inputs into varied follow-up inputs. These inputs are processed by the system under test (SUT), and the outputs are scrutinized to determine the model's adaptability and accuracy under diverse conditions. The framework's key components include:

\begin{itemize}
    \item \textbf{Metamorphic Relation Creation:} This component defines MRs with specific criteria for generating and processing test cases, ensuring each test is capable of effectively evaluating the model’s capabilities.
    \item \textbf{Transformation Function:} Alters source inputs to create diverse follow-up inputs, challenging the model’s capacity to handle a range of input scenarios.
    \item \textbf{Relation Function:} Assesses the consistency of outputs from the SUT against expected outcomes defined by the MRs, gauging adaptability and accuracy.
    \item \textbf{System Under Test:} Refers to the model or algorithm being evaluated, which processes inputs and produces outputs. Its performance and robustness are critically examined through these tests.
\end{itemize}

Please note that the detailed framework and methods, including Figure \ref{fig:metamorphic_framework}, are part of a closed-source development and are not currently publicly available.

By leveraging the GeMTest framework, we ensure robust evaluation of models against both expected and complex input scenarios, significantly enhancing their reliability for real-world applications. This framework also allows for further advancements, such as retraining models based on test results to continually improve their adaptability and effectiveness.

\subsection{Learning Algorithms}

\subsubsection{Supervised Learning}
Supervised learning is the most prevalent form of ML, where the model learns from labeled training data to predict outcomes. The training involves input-output pairs where the desired outputs (labels) guide the learning process. However, this method often requires a substantial amount of labeled data, which can be costly and time-consuming to obtain. Additionally, supervised learning models are prone to overfitting, especially when the labeled data does not represent the full spectrum of real-world scenarios. This limitation becomes evident when models trained on limited data fail to generalize well outside their training set \cite{Vermeulen2020}.

\subsubsection{Semi-Supervised Learning}

Given the challenges associated with supervised learning, especially in contexts requiring robust and generalized models, semi-supervised learning presents a compelling alternative. Semi-supervised learning leverages both labeled and unlabeled data, making it particularly suitable for applications where labeled data is scarce but unlabeled data is abundant. This approach is more resource-efficient and reduces the risk of overfitting by utilizing a broader dataset that more closely represents real-world distributions.

The choice of semi-supervised learning in our metamorphic testing framework is strategic. The framework itself does not rely on ground truth labels for validating the model outputs, which aligns perfectly with the semi-supervised paradigm where pseudo labels generated from unlabeled data can be effectively utilized. By incorporating unlabeled data into the training process, semi-supervised learning enriches the training environment and exposes the model to a wider variety of input scenarios. This exposure is crucial for enhancing the model's robustness, enabling it to perform reliably in unexpected real-world conditions \cite{NEURIPS2022_15dce910}. 

Moreover, semi-supervised learning methods facilitate the integration of metamorphic testing by allowing the seamless inclusion of transformed data (generated through metamorphic relations) into the training process. This integration helps continuously refine the model's ability to generalize from both labeled and pseudo-labeled data, thus improving the overall robustness and accuracy of the system under diverse operational scenarios.
\newline
\newline
In summary, semi-supervised learning not only addresses the limitations of supervised learning by making efficient use of available data but also enhances the adaptability and robustness of models in practical applications. It supports a dynamic learning environment where models can be iteratively tested and refined within the metamorphic testing framework, ensuring their reliability and effectiveness across varied conditions.

\subsection{Semi-Supervised Algorithms}

In this section, we describe the four semi-supervised learning algorithms we selected that utilize both labeled and unlabeled data to improve the performance of the model, each method being unique to effectively balance the learning process.

\subsubsection{\textbf{FixMatch}}
\begin{figure}[h]  
    \centering  
    \includegraphics[width=0.5\textwidth]{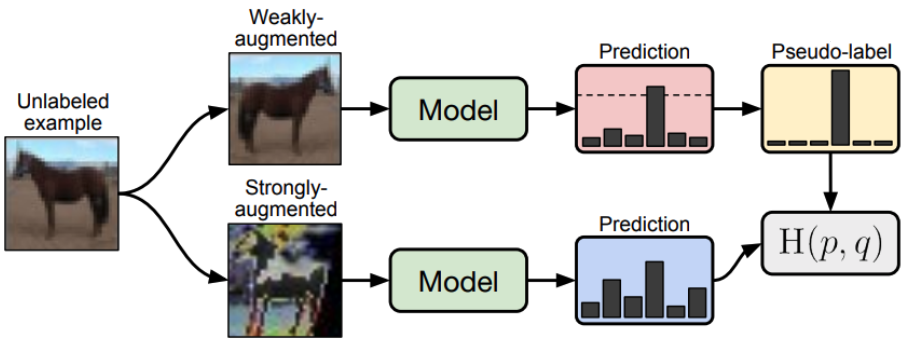}  %
    \caption{Architecture of FixMatch \cite{200107685}}  
    \label{fig:example-image}  
\end{figure}

FixMatch simplifies the use of pseudo-labels by applying consistency regularization between weakly and strongly augmented versions of the same data. It enforces high-confidence predictions for unlabeled data to be treated as labels, guided by a static confidence threshold:
\[
L = L_{\text{sup}} + \lambda_u \cdot L_{\text{unsup}}
\]
Where \( L_{\text{sup}} \) represents the loss on labeled data and \( L_{\text{unsup}} \) is the loss on pseudo-labeled data, scaled by a weighting factor \( \lambda_u \). The pseudo-labels are generated from the weakly augmented data if the model's confidence exceeds a predefined threshold \cite{200107685}.
\subsubsection{\textbf{FlexMatch}}
\begin{figure}[h]  
    \centering  
    \includegraphics[width=0.5\textwidth]{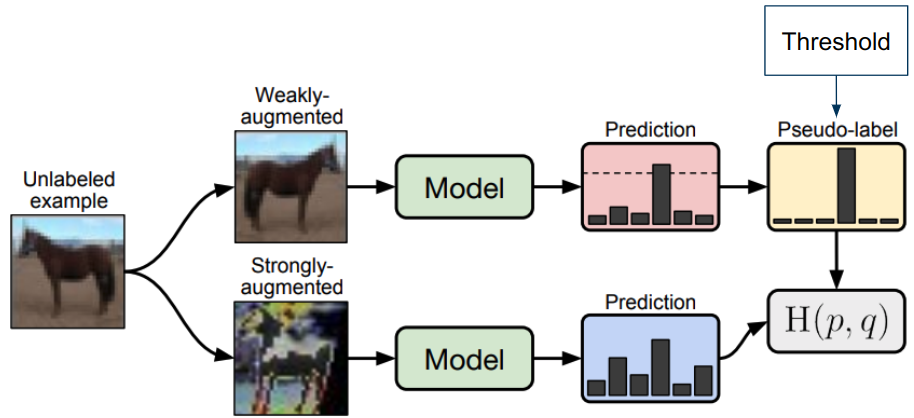}  %
    \caption{Architecture of FlexMatch \cite{200107685}}  
    \label{fig:example-image}  
\end{figure}
FlexMatch extends FixMatch by dynamically adjusting the confidence thresholds based on the class-wise difficulty, thus addressing the class imbalance:
\[
L = L_{\text{sup}} + \lambda_u \cdot \sum_{c=1}^C \tau_c L_{\text{unsup}, c}
\]
where \( C \) is the number of classes, \( \tau_c \) is the dynamic threshold for class \( c \), and \( L_{\text{unsup}, c} \) is the unsupervised loss for class \( c \). FlexMatch thereby allows for more adaptive learning across varying class distributions \cite{211008263}.
\subsubsection{\textbf{MixMatch}}
MixMatch introduces an algorithm that averages the labels and mixes up the inputs and labels, utilizing the MixUp technique:
\[
x' = \gamma x_i + (1 - \gamma) x_j, \quad y' = \gamma y_i + (1 - \gamma) y_j
\]
Here, \( (x_i, y_i) \) and \( (x_j, y_j) \) are input-label pairs, mixed by coefficient \( \gamma \), which is typically sampled from a Beta distribution. MixMatch thus encourages the model to learn from a smoother distribution of data \cite{90502249}.
\subsubsection{\textbf{FullMatch}}
FullMatch further develops the ideas presented in MixMatch and FixMatch by incorporating adaptive penalty mechanisms to penalize incorrect pseudo-labels more effectively:
\[
L = L_{\text{sup}} + \lambda_u \cdot L_{\text{unsup}} + \lambda_p \cdot L_{\text{penalty}}
\]
Where \( L_{\text{penalty}} \) involves penalties such as entropy minimization and adaptive negative learning, ensuring that the pseudo-labels do not reinforce incorrect behaviors. This makes FullMatch suitable for complex and highly varied data scenarios, where robustness against noisy labels is crucial \cite{chen2023boostingsemisupervisedlearningexploiting}.
\newline
\newline
Each algorithm utilizes the concept of learning from both labeled and unlabeled data but introduces different techniques to optimize the learning process. FixMatch focuses on a simple yet effective threshold-based approach, FlexMatch adapts thresholds dynamically, MixMatch leverages data augmentation to enhance generalization, and FullMatch introduces sophisticated penalty mechanisms to mitigate the impact of noisy labels. Together, these algorithms form a versatile toolkit for semi-supervised learning, allowing for effective empirical comparisons across different scenarios.

\subsection{Datasets}

To evaluate the performance of our semi-supervised learning algorithms, we utilized three widely recognized benchmark datasets, each offering unique challenges and characteristics.

\subsubsection{\textbf{MNIST}}
The MNIST dataset consists of handwritten digits (0 through 9) and contains 60,000 training images and 10,000 testing images. Each image is a 28x28 pixel grayscale representation of a digit. We chose MNIST due to its simplicity and effectiveness in benchmarking image classification algorithms, providing a straightforward dataset where model improvements can be easily quantified \cite{deng2012mnist}.

\subsubsection{\textbf{CIFAR-10}}
The CIFAR-10 dataset includes 60,000 32x32 color images in 10 different classes, with 6,000 images per class. The dataset is split into 50,000 training images and 10,000 testing images. CIFAR-10 is significantly more complex than MNIST due to its color images and more varied backgrounds, making it a suitable choice for evaluating the robustness and effectiveness of our learning models under more challenging conditions \cite{krizhevsky2009learning}.

\subsubsection{\textbf{CIFAR-100}}
Similar to CIFAR-10, the CIFAR-100 dataset features 32x32 color images. However, it contains 100 classes with 600 images each (500 training images and 100 testing images per class). The increased number of classes introduces a higher level of granularity, making CIFAR-100 ideal for testing the algorithms' capability to differentiate between a larger variety of objects, thus providing insights into the scalability and precision of our methods \cite{krizhevsky2009learning}.

\subsection{Models}

The models selected for this study are standard architectures in computer vision, known for their robustness and widespread use across various image recognition tasks. Each model brings different architectural benefits, making them suitable for our comparative analysis.

\subsubsection{\textbf{ResNet-32}}
ResNet-32 is a variant of the Residual Network architecture that includes 32 layers. It is well-known for its ability to combat the vanishing gradient problem through skip connections that allow for deeper networks. This model is particularly useful for our studies on CIFAR-10 and CIFAR-100 due to its balance between depth and computational efficiency \cite{he2015deepresiduallearningimage}.

\subsubsection{\textbf{ResNet-50}}
ResNet-50, a deeper variant with 50 layers, is utilized for its enhanced capacity for learning from complex image data sets like CIFAR-100. The additional layers allow for more abstract feature representations, making it highly effective for detailed image classification tasks \cite{he2015deepresiduallearningimage}.

\subsubsection{\textbf{VGG16}}
VGG16 is renowned for its simplicity and depth, consisting of 16 convolutional layers. It is excellent for feature extraction in image processing tasks. We included VGG16 to benchmark its performance against residual architectures and to take advantage of its robustness in handling more textured and varied image backgrounds found in datasets like CIFAR-10 \cite{SimonyanZ14a}.

\subsubsection{\textbf{FCN}}
Fully Convolutional Networks (FCN) are pivotal in tasks that involve spatial data preservation such as semantic segmentation. We included FCN to explore how semi-supervised learning adaptations might benefit tasks beyond simple classification, leveraging its architecture to examine spatial hierarchies and contextual dependencies in images \cite{7298965}.
\newline
\newline
The combination of MNIST, CIFAR-10, and CIFAR-100 datasets along with ResNet-32, ResNet-50, VGG16, and FCN models provides a comprehensive platform to evaluate the performance of semi-supervised learning algorithms across various levels of image complexity and task requirements. This diverse set of datasets and models helps in assessing the generalizability and scalability of the algorithms, ensuring that findings are robust across different scenarios and architectural challenges.

\section{Methodology}
\begin{figure*}[t]
    \centering
    \includegraphics[width=\textwidth]{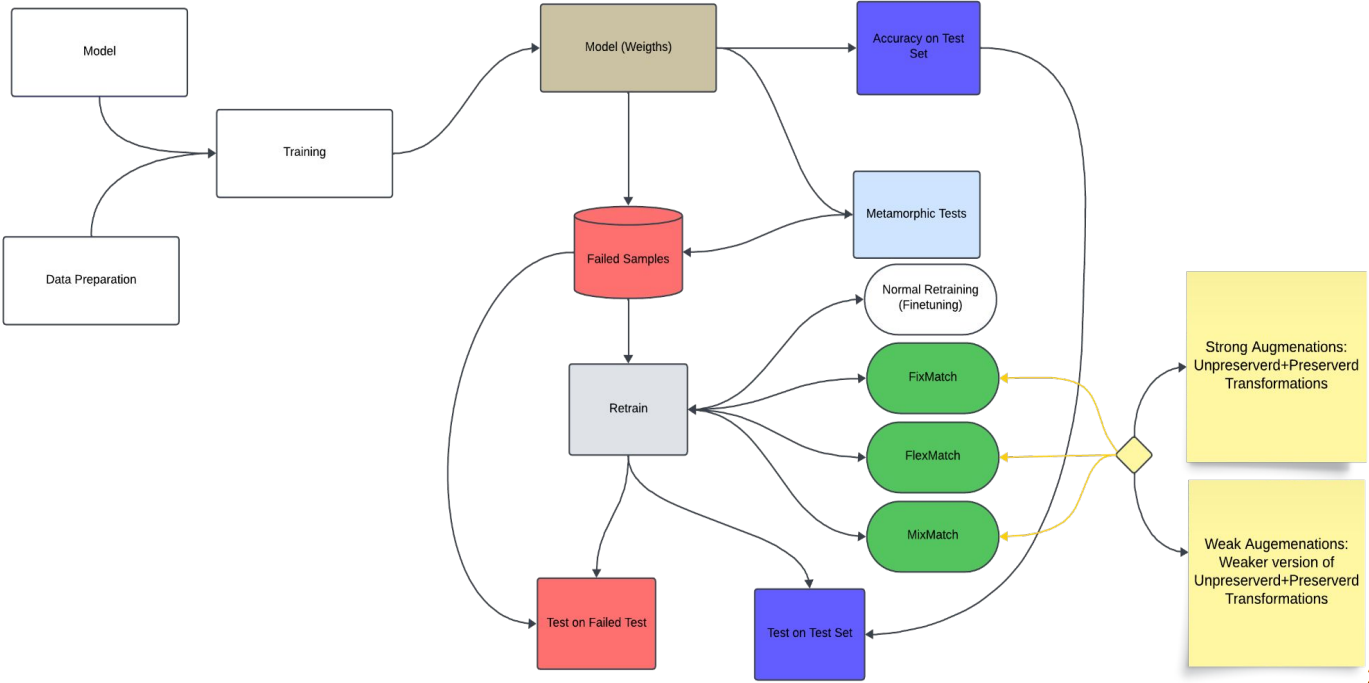}
    \caption{Architecture of our proposed pipeline}
    \label{fig:pipeline}
\end{figure*}
\subsection{Pipeline}

\subsubsection{Overview}
The key idea of our pipeline is to apply a repetitive feedback loop of Retraining with a data set augmented with the target metamorphic relations, evaluate the model, and test again on metamorphic tests formed from the available pool of metamorphic relations, and then extract the metamorphic relations from the failed and passed tests, set the target and metamorphic relations and repeat.

\subsubsection{Metamorphic Tester}
The first component of our pipeline is the metamorphic tester which instantiates the models and the datasets prepares the metamorphic relations into the correct format, and forms test suits using the GeMTest framework for metamorphic testing that takes test images and augments them with a metamorphic relation, giving back the augmented images and the new target labels and using this generated set it runs the images through the model and compares the predicted label with the new target label and based on the success rate the test is deemed passed or failed.

The model is tested with the test suits, and following that, we form two sets of test suits based on which passed and which didn't, and from these sets, we extract the metamorphic relations themselves as functions to be used later in the cycle when retraining.

Using the extracted metamorphic relations, we generate a new data set and form an augmented PyTorch data loader from it that we use for the retraining process.

\subsubsection{Model Retraining}

With the augmented training set we can then use any of the provided semi-supervised algorithms implementation to retrain our model or use the base retraining process, after choosing how the retraining will happen, we can then choose how will the data be augmented for the next cycle, for this part we have one of two options, the first being to retrain adaptively based on the extracted failed tests allowing for a more responsive approach that takes into considerations the model's weaknesses based on the failed metamorphic tests, and the second option to generate the new set with a fixed ratios of all the metamorphic relations.

We can conduct iterative cycles by either setting a fixed number of iterations or establishing a stopping criterion. This criterion may involve halting the process if the model's performance falls below or exceeds a certain threshold. We employ different retraining methods, which include:

Retraining with the base semi-supervised algorithms using their standard data augmentations.
\begin{tcolorbox}[colback=gray!20, colframe=black, title=Retraining Techniques]
\begin{itemize}
    \item \textbf{Base Algorithm}: Retraining with the base semi-supervised algorithms using their standard data augmentations. 
    
    \item \textbf{The Adaptive Method}:  which specifies the failed tests as strong augmentations for the semi-supervised algorithms.
    
    \item \textbf{The Static Method}: A mode where weak augmentations consist of single metamorphic relations and strong augmentations are combinations of these relations.
\end{itemize}
\end{tcolorbox}

In the results section, we compare the performance of each of these modes across the previously mentioned models.

We also use ONNX \cite{onnxruntime}, which is an open, highly compatible format built to represent ML models. Where it defines a common set of operators - the building blocks of ML and DL models - and a common file format to enable us to use models with a variety of frameworks, tools, runtimes, and compilers to ensure that the loaded models can be used exported from and to any other ML framework for the maximum possible compatibility.

Moreover, in our ML workflow, the retraining and evaluation pipeline is designed for parallel execution, utilizing separate threads for each task to maximize efficiency. This parallelizable approach leverages multi-threading to handle different stages of the pipeline concurrently, reducing overall processing time and optimizing resource usage. Specifically, data preprocessing, model training, and evaluation are each allocated distinct threads, enabling simultaneous execution without bottlenecks. By employing this method, we can ensure that the system scales effectively with increased data volume and complexity, maintaining robust performance and expediting the iterative process of model refinement. This design not only enhances throughput but also improves the responsiveness and agility of our ML operations, making it well-suited for dynamic and high-demand environments.

Lastly, the preparation and caching of resources, such as MRs and augmented sets (new data loaders), for the next iteration are efficiently parallelized. By utilizing separate threads, we can concurrently test the model while training it.

\section{Experimental Setup}

In this section, we provide detailed insights into the methodologies and setups employed to validate our hypotheses. This section encompasses a comprehensive explanation of the evaluation metrics used, a thorough description of the datasets and preprocessing techniques, and an in-depth analysis of the models, including their architectures, training processes, and any modifications made.

\subsection{Metrics}

\subsubsection{Robustness}

Robustness is a critical metric for evaluating ML models, particularly in real-world applications where input data can vary significantly. According to \cite{braiek2024machinelearningrobustnessprimer}, robustness denotes the capacity of a model to sustain stable predictive performance in the face of variations and changes in the input data. This definition underscores the importance of a model's resilience to changes, ensuring that it can maintain high performance even when the input data is perturbed.

In our context, robustness is defined below as inspired by \cite{braiek2024machinelearningrobustnessprimer}
\begin{tcolorbox}[colback=gray!20, colframe=black, title=Definition: Robustness]
Robustness refers to a model’s ability to function properly on noisy or otherwise perturbed data \cite{9000651}.
\end{tcolorbox}

A significant advantage of this definition of robustness is that it can be applied to unlabeled data as well. This makes it an excellent addition in the context of semi-supervised learning, where labeled data is scarce, and the model needs to generalize well to new, unseen, and unlabeled data.

To quantify robustness, we use the success rate on metamorphic tests. The success rate on metamorphic tests evaluates the model's consistency.

\begin{equation}
SR_{MT} = \frac{1}{N} \sum_{i=1}^{N} MTest(x_i, M)
\end{equation}

Where:
\begin{itemize}
    \item \(N\) is the total number of metamorphic test cases.
    \item \(x_i\) represents the \(i\)-th input test data.
    \item \(M\) is the model (system under test).
    \item \(MTest(x_i, M)\) is a function that returns 1 if the model \(M\) produces a consistent or correct output for the input \(x_i\) after applying metamorphic transformations, and 0 otherwise.
\end{itemize}

While robustness is a crucial metric for evaluating a model's resilience to input variations, relying solely on robustness can be misleading. A model that consistently outputs the same value for all inputs can achieve perfect robustness because its output does not change regardless of input perturbations. However, such a model is practically useless because it lacks the ability to make accurate predictions based on the input data.

For instance, consider a hypothetical figure illustrating the robustness of a model that always outputs the same value. This model would have a robustness score of 1 (or 100\%) on metamorphic tests because its output remains invariant under any transformation. However, this invariant output fails to provide any meaningful information or correct predictions, rendering the model ineffective for practical use.

Therefore, it is essential to evaluate robustness in conjunction with accuracy. Accuracy measures the proportion of correct predictions made by the model, providing a direct assessment of its predictive performance. By considering both robustness and accuracy, we can ensure that the model is not only resilient to variations in input data but also capable of making accurate predictions. This dual evaluation helps in developing models that are both reliable and effective in real-world applications.

\begin{figure}[h]  
    \centering  
    \includegraphics[width=0.5\textwidth]{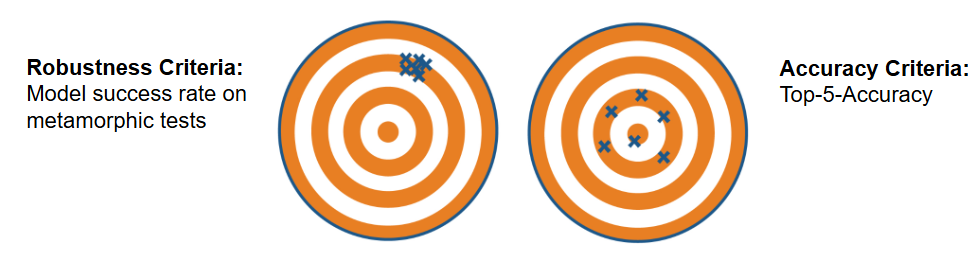}  
    \caption{Robustness and accuracy.}  
    \label{fig:example-image}  
\end{figure}

\subsubsection{Accuracy}

Accuracy is a fundamental metric for evaluating the performance of a machine-learning model. In our experiments, we use \textbf{top-\(\mathcal{N}\) accuracy} on the test set as a key performance indicator. This metric is particularly useful for classification tasks where the goal is to measure how well the model's predictions align with the true labels.

Top-\(\mathcal{N}\) accuracy measures the proportion of test samples for which the true label is among the top \(\mathcal{N}\) predicted labels by the model. It provides an indication of how often the correct label appears within the top \(\mathcal{N}\) predictions made by the model.

Formally, top-5 accuracy (\(Acc_{top5}\)) is defined as:

\begin{equation}
Acc_{top5} = \frac{1}{M} \sum_{j=1}^{M} \mathbb{I}\left( y_j \in \text{Top-\(\mathcal{N}\) Predictions}(x_j) \right)
\end{equation}

Where:
\begin{itemize}
    \item \(M\) is the total number of test samples.
    \item \(x_j\) represents the \(j\)-th test input.
    \item \(y_j\) is the true label for \(x_j\).
    \item \(\text{Top-\(\mathcal{N}\) Predictions}(x_j)\) refers to the top 5 predicted labels for input \(x_j\).
    \item \(\mathbb{I}(\cdot)\) is an indicator function that returns 1 if the condition is true (i.e., the true label \(y_j\) is within the top 5 predictions) and 0 otherwise.
\end{itemize}

The test set used for evaluating top-\(\mathcal{N}\) accuracy consists of new and unseen data, which is critical for assessing the model's generalization capabilities. By evaluating the model on data that was not used during training, we gauge how well the model can adapt to new examples and whether it is effectively leveraging the labeled data available.

In the context of semi-supervised learning, where only a small subset of the data is labeled, top-\(\mathcal{N}\) accuracy serves as a useful metric to measure how well the model is adapting and performing with the limited labeled set. It provides insight into the model’s ability to generalize from the small labeled data while also leveraging the large pool of unlabeled data to make accurate predictions.

\subsection{Base Methods vs Adaptive Retraining}

\subsubsection{Objective}

This experiment aims to compare the performance of base semi-supervised algorithm methods against the same methods enhanced with an adaptive retraining schema that we have implemented. The primary focus is to evaluate how each configuration handles the trade-off between accuracy and robustness. We use two key metrics for this comparison: \textit{accuracy} and \textit{robustness}. This will help determine which method provides a better balance between these two critical aspects.

\subsubsection{Experimental Setup}
\begin{itemize}
    \item \textbf{Base Configuration}: Utilizes the standard version of the semi-supervised algorithms without any additional retraining mechanisms. This setup provides a baseline for comparison.
    \item \textbf{Adaptive Retraining Configuration}: Incorporates an adaptive retraining schema that dynamically adjusts parameters based on performance metrics during training. This approach aims to enhance the robustness of the algorithm.
\end{itemize}

We present results from two different runs. Each run used the same dataset for all four semi-supervised learning algorithms and their adaptive retraining counterparts. Accuracy was measured as the top-5 accuracy, as detailed above. Robustness was calculated using the Success Rate on metamorphic tests. Both runs used 0.1\% of the original dataset to simulate data scarcity. This subset was then split into 10\% for the labeled set, 70\% for the unlabeled set, and 20\% for the test set. Each training loop began with an untrained model and ran for 30 Robustness cycles, with each cycle consisting of two epochs on the same data. The model used was ResNet-50.

We also utilized Robustness cycles. In each Robustness cycle, a new adaptive or static dataset was produced.

\begin{tcolorbox}[colback=white, colframe=black, title=Experimental Configuration]
\begin{itemize}
    \item \textbf{Dataset Size}: 0.1\% of original dataset 
    \item \textbf{Dataset}: CIFAR-10, CIFAR-100
    \item \textbf{Data Split}: 10\% labeled, 70\% unlabeled, 20\% test
    \item \textbf{Algorithms}: 4 semi-supervised + adaptive retraining counterparts
    \item \textbf{Accuracy Metric}: Top-5 accuracy
    \item \textbf{Robustness Metric}: Success rate on Metamorphic Tests
    \item \textbf{Model}: ResNet-50
    \item \textbf{Training Cycles}: 30 robustness cycles, each with 2 epochs over the entire dataset
    \item \textbf{Initial Condition}: Untrained model
\end{itemize}
\end{tcolorbox}

\subsubsection{Results}
The results of this experiment are illustrated in \ref{tab:Base-method-vs-adaptive-Retraining}, which shows the comparative performance of the base and adaptive retraining configurations across both metrics. The \ref{tab:Base-method-vs-adaptive-Retraining-CIFAR-10} shows the results on a much more difficult task. The same data patterns can be seen with much smaller margins because the task of getting top-5 accuracy is exponentially harder when you have ten times more classes.

1. \textbf{Accuracy vs. Robustness}: The data indicates that metamorphic retraining, on average, provides better robustness compared to the Base algorithms. However, Base algorithms demonstrate superior accuracy.

2. \textbf{Non-linear Trade-offs}: The trade-off between accuracy and robustness is not linear. For instance, the Base FixMatch algorithm achieves the second-best performance in both accuracy and robustness among the algorithms tested. This highlights that improvements in one metric do not necessarily mean proportional improvements in the other.

3. \textbf{Flexmatch Performance}: The Base Flexmatch algorithm shows high robustness relative to other methods. This observation is attributed to the dynamic thresholding employed by Flexmatch, which allows it to adapt more effectively to failed instances by adjusting its thresholds based on ongoing performance. This adaptability contributes to its robustness, which can match adaptive retraining Flexmatch.

4. \textbf{FullMatch Analysis}: Although the base FullMatch algorithm does not perform exceptionally well, the FullMatch with adaptive retraining demonstrates the best trade-off between robustness and accuracy as it got second place in both robustness and accuracy. The adaptive retraining version of FullMatch maintains robustness more consistently throughout training, indicating that adaptive mechanisms can effectively balance the trade-off between accuracy and robustness.

In conclusion, while base methods generally provide higher accuracy, adaptive retraining methods, particularly FullMatch with adaptive retraining, offer a better balance of robustness. This experiment underscores the importance of incorporating adaptive strategies to achieve a more stable and reliable performance in semi-supervised learning algorithms.

\begin{table}[h]
    \centering
    \begin{tabular}{|>{\centering\arraybackslash}m{2cm}|c|c|c|c|}
        \hline
        \rowcolor{white}
        & \multicolumn{2}{c|}{\textbf{Base Method}} & \multicolumn{2}{c|}{\textbf{Adaptive Retraining}} \\
        \cline{2-5}
        \rowcolor{white}
        & \textbf{Accuracy} & \textbf{Robustness} & \textbf{Accuracy} & \textbf{Robustness} \\
        \hline
        \textbf{FixMatch} & 70\% & 75\% & 55\% & 84\% \\
        \hline
        \textbf{FlexMatch} & 60\% & 90\% & 60\% & 83\% \\
        \hline
        \textbf{MixMatch} & 65\% & 70\% & 70\% & 85\% \\
        \hline
        \textbf{FullMatch} & 75\% & 68\% & 68\% & 87\% \\
        \hline
        \rowcolor[gray]{.9}
        \textbf{Average} & \cellcolor[HTML]{90EE90}67.5\% & 75.75\% & 63.25\% & \cellcolor[HTML]{90EE90}84.75\% \\
        \hline
    \end{tabular}
    \vspace{0.1cm}
    \caption{Comparison of accuracy and robustness between base method and adaptive retraining for \textbf{CIFAR-10}}
    \label{tab:Base-method-vs-adaptive-Retraining}
\end{table}

\begin{table}[h]
    \centering
    \begin{tabular}{|>{\centering\arraybackslash}m{2cm}|c|c|c|c|}
        \hline
        \rowcolor{white}
        & \multicolumn{2}{c|}{\textbf{Base Method}} & \multicolumn{2}{c|}{\textbf{Adaptive Retraining}} \\
        \cline{2-5}
        \rowcolor{white}
        & \textbf{Accuracy} & \textbf{Robustness} & \textbf{Accuracy} & \textbf{Robustness} \\
        \hline
        \textbf{FixMatch} & 14\% & 79\% & 15\% & 57\% \\
        \hline
        \textbf{FlexMatch} & 6\% & 100\% & 6\% & 80\% \\
        \hline
        \textbf{MixMatch} & 6\% & 58\% & 16\% & 90\% \\
        \hline
        \textbf{FullMatch} & 22\% & 53\% & 7\% & 63\% \\
        \hline
        \rowcolor[gray]{.9}
        \textbf{Average} & \cellcolor[HTML]{90EE90}12\% & 72.5\% & 11\% & \cellcolor[HTML]{90EE90}72.5\% \\
        \hline
    \end{tabular}
    \vspace{0.1cm}
    \caption{Comparison of accuracy and robustness between base method and adaptive retraining for \textbf{CIFAR-100}}
    \label{tab:Base-method-vs-adaptive-Retraining-CIFAR-10}
\end{table}

\subsection{Impact on Pretrained Models}

\subsubsection{Objective}
The goal of this experiment is to evaluate the effect of adaptive retraining on pretrained models in comparison to non-pretrained models. We aim to determine whether pretrained models exhibit improved robustness and accuracy when subjected to our adaptive retraining schema. By focusing on both accuracy and robustness metrics, we seek to understand if pretrained models offer a better trade-off between these two critical performance measures.

\subsubsection{Experimental Setup}
This experiment involves two primary configurations for each algorithm: pretrained and non-pretrained models. The pretrained models have undergone initial training on a large, relevant dataset, providing a solid foundation for further refinement. In contrast, the non-pretrained models start from scratch without any prior knowledge. The pretrained model had the last four layers set to trainable; they ran for 30 robustness cycles with two epochs per cycle. The model was VGG16.

\begin{tcolorbox}[colback=white, colframe=black, title=Experimental Configuration]
\begin{itemize}
    \item \textbf{Dataset Size}: 0.1\% of original dataset 
    \item \textbf{Dataset}: CIFAR-10, MNIST
    \item \textbf{Data Split}: 10\% labeled, 70\% unlabeled, 20\% test
    \item \textbf{Algorithms}: FixMatch, FullMatch, Flexmatch, Mixmatch, all with adaptive retraining
    \item \textbf{Configurations}: Pretrained, Non-pretrained
    \item \textbf{Accuracy Metric}: Top-5 accuracy
    \item \textbf{Robustness Metric}: Success rate on Metamorphic Tests
    \item \textbf{Model}: VGG16
    \item \textbf{Pretrained Layers}: Last 4 layers set to trainable
    \item \textbf{Training Cycles}: 30 Robustness cycles with each cycle 2 epochs
\end{itemize}
\end{tcolorbox}

\begin{itemize}
    \item \textbf{Pretrained Models with Adaptive Retraining}: These models leverage the adaptive retraining schema to fine-tune the already established pretrained parameters. The aim is to enhance both accuracy and robustness by building on the solid foundation provided by the pretrained models.
    \item \textbf{Non-Pretrained Models with Adaptive Retraining}: These models utilize the adaptive retraining schema from an untrained state. This setup helps to highlight the relative benefits of starting with pretrained parameters.
\end{itemize}

\begin{table}[h]
    \centering
    \begin{tabular}{|>{\centering\arraybackslash}m{2cm}|c|c|c|c|}
        \hline
        \rowcolor{white}
        & \multicolumn{2}{c|}{\textbf{Random Initialization}} & \multicolumn{2}{c|}{\textbf{Pretrained}} \\
        \cline{2-5}
        \rowcolor{white}
        & \textbf{Accuracy} & \textbf{Robustness} & \textbf{Accuracy} & \textbf{Robustness} \\
        \hline
        \textbf{FixMatch} & 75\% & 50\% & 90\% & 80\% \\
        \hline
        \textbf{FlexMatch} & 67\% & 70\% & 90\% & 72\% \\
        \hline
        \textbf{MixMatch} & 67\% & 70\% & 70\% & 73\% \\
        \hline
        \textbf{FullMatch} & 70\% & 63\% & 80\% & 70\% \\
        \hline
        \rowcolor[gray]{.9}
        \textbf{Average} & 69.5\% & 63.25\% & \cellcolor[HTML]{90EE90}82.5\% & \cellcolor[HTML]{90EE90}73.75\% \\
        \hline
    \end{tabular}
    \vspace{0.3cm}
    \caption{Impact on Pretrained Models}
    \label{tab:impact-pretrained}
\end{table}

\subsubsection{Results}
The results from this experiment are displayed in \ref{tab:impact-pretrained}, which compares the performance of pretrained and non-pretrained models under adaptive retraining.

\begin{itemize}
    \item \textbf{Pretrained Models Performance}: Pretrained models demonstrate significantly better performance in terms of both accuracy and robustness when subjected to adaptive retraining. This improvement suggests that starting with a strong, pretrained foundation allows the adaptive retraining schema to fine-tune the model more effectively.
    \item \textbf{Accuracy and Robustness Trade-off}: Pretrained models not only show higher accuracy, which is expected due to their initial training, but also exhibit enhanced robustness. This indicates a better trade-off between accuracy and robustness compared to non-pretrained models. The results suggest that pretrained models provide a more stable baseline, allowing adaptive mechanisms to optimize performance more effectively.
\end{itemize}

In summary, this experiment underscores the advantages of using pretrained models with adaptive retraining. The combination of a strong initial foundation and adaptive mechanisms leads to superior performance in terms of both accuracy and robustness. FixMatch, in particular, excels under these conditions. This shows the capability of our method to robustify pretrained models without requiring full retraining.

\subsection{Best Configuration for Non-Scarce Data}

\subsubsection{Objective}
The purpose of this experiment is to evaluate the performance of various semi-supervised learning algorithms in scenarios where there is an abundance of labeled data. Specifically, we aim to assess the impact of adaptive retraining on both pretrained and non-pretrained models compared to base algorithms. This experiment explores whether the robustification and accuracy benefits of our method persist when data is plentiful and whether simpler approaches might suffice under these conditions.

\subsubsection{Experimental Setup}
For this experiment, all algorithms were tested under conditions of abundant labeled data. The setups included both pretrained and non-pretrained models, with and without adaptive retraining. The algorithms tested were FixMatch and FullMatch, alongside their base versions.

\begin{itemize}
    \item \textbf{Pretrained Models with Adaptive Retraining}: Models initially trained on a large, relevant dataset, followed by adaptive retraining.
    \item \textbf{Non-Pretrained Models with Adaptive Retraining}: Models starting from scratch, with adaptive retraining applied.
    \item \textbf{Base Algorithms}: Both pretrained and non-pretrained models without adaptive retraining.
\end{itemize}

The primary metrics evaluated were accuracy and robustness, with the expectation that an abundance of labeled data would allow models to achieve high performance with minimal need for advanced techniques.

\begin{tcolorbox}[colback=white, colframe=black, title=Experimental Configuration]
\begin{itemize}
    \item \textbf{Dataset Size}: 90\% labeled set and 10\% test set
    \item \textbf{Dataset}: CIFAR-100
    \item \textbf{Configurations}: Pretrained, Non-pretrained, with and without adaptive retraining
    \item \textbf{Algorithms}: FixMatch, FullMatch, Flexmatch, MixMatch. Base and adaptive retraining version.
    \item \textbf{Accuracy Metric}: Top-5 accuracy
    \item \textbf{Robustness Metric}: Success rate on Metamorphic Tests
    \item \textbf{Model}: ResNet 50
    \item \textbf{Pretrained Layers}: Last 4 layers set to trainable
    \item \textbf{Training Cycles}: 30 Robustness cycles
    \item \textbf{Epochs per Cycle}: 1 epoch
\end{itemize}
\end{tcolorbox}

\subsubsection{Results}
Table \ref{tab:large-dataset} illustrates the performance of all configurations under conditions of abundant labeled data.

\begin{itemize}
    \item \textbf{Loss Fluctuation and Convergence}: The results show that the loss fluctuates with each robustification cycle due to changes in the data loader transformers. However, all algorithms eventually converge, indicating that the models can learn effectively from the ample labeled dataset. This convergence is expected since the abundance of data reduces the reliance on semi-supervised learning techniques.
    \item \textbf{State-of-the-Art Accuracy and Robustification}: With sufficient labeled data, all algorithms, whether pretrained or non-pretrained, with or without adaptive retraining, achieve state-of-the-art accuracy and perfect robustification. This demonstrates that, under these conditions, the models can fully leverage the labeled data to achieve optimal performance.
    \item \textbf{Comparison Across Configurations}:
    \begin{itemize}
        \item \textbf{Pretrained with Adaptive Retraining}: These models continue to show excellent performance, but the gap between them and simpler configurations narrows significantly when data is abundant.
        \item \textbf{Non-Pretrained with Adaptive Retraining}: These models also perform exceptionally well, benefiting greatly from the rich labeled dataset.
        \item \textbf{Base Algorithms}: Interestingly, the base algorithms, both pretrained and non-pretrained, achieve comparable performance to their adaptively retrained counterparts. This suggests that the abundance of data mitigates the need for advanced retraining techniques.
    \end{itemize}
    \item \textbf{Implications}: Table \ref{tab:large-dataset} suggests that for tasks with abundant labeled data, any configuration of the algorithms will likely perform well. This includes base versions without adaptive retraining. All methods achieved very similar stats, but there was no clear pattern. The only thing worth noting is that FullMatch with adaptive retraining seems to be quite better than the vanilla implementation. Simpler approaches, such as regular data augmentations, may be more efficient and equally effective for such tasks. This finding indicates that the adaptive retraining framework may be better suited for scenarios with limited labeled data, where its benefits are more pronounced.
\end{itemize}

In conclusion, this experiment highlights that while our adaptive retraining framework offers significant advantages in data-scarce environments, its benefits diminish when labeled data is plentiful. For such tasks, simpler approaches may be more practical and efficient, underscoring the importance of choosing the right tool for the right problem.

\begin{table}[h]
    \centering
    \begin{tabular}{|>{\centering\arraybackslash}m{2cm}|c|c|c|c|}
        \hline
        \rowcolor{white}
        & \multicolumn{2}{c|}{\textbf{Base Method}} & \multicolumn{2}{c|}{\textbf{Adaptive Retraining}} \\
        \cline{2-5}
        \rowcolor{white}
        & \textbf{Accuracy} & \textbf{Robustness} & \textbf{Accuracy} & \textbf{Robustness} \\
        \hline
        \textbf{FixMatch} & 22\% & 100\% & 24\% & 100\% \\
        \hline
        \textbf{FlexMatch} & 28\% & 100\% & 27\% & 100\% \\
        \hline
        \textbf{MixMatch} & 26\% & 100\% & 26\% & 100\% \\
        \hline
        \textbf{FullMatch} & 23\% & 100\% & 28\% & 100\% \\
        \hline
        \rowcolor[gray]{.9}
        \textbf{Average} & 24.75\% & 100\% & 26.25\% & 100\% \\
        \hline
    \end{tabular}
    \vspace{0.3cm}
    \caption{Performance on large labeled dataset}
    \label{tab:large-dataset}
\end{table}

\subsection{Experiment 4: Performance While Taking into Account Non-Label Preserving Transformations}

\subsubsection{Objective}
This experiment aims to investigate the performance of different semi-supervised learning algorithms when subjected to non-label preserving transformations. Specifically, we examine how well these algorithms handle metamorphic tests that alter both the input and output labels. The focus is on robustness and accuracy in the face of such transformations.

\subsubsection{Experimental Setup}
\begin{figure}[h]  
    \centering  
    \includegraphics[width=0.5\textwidth]{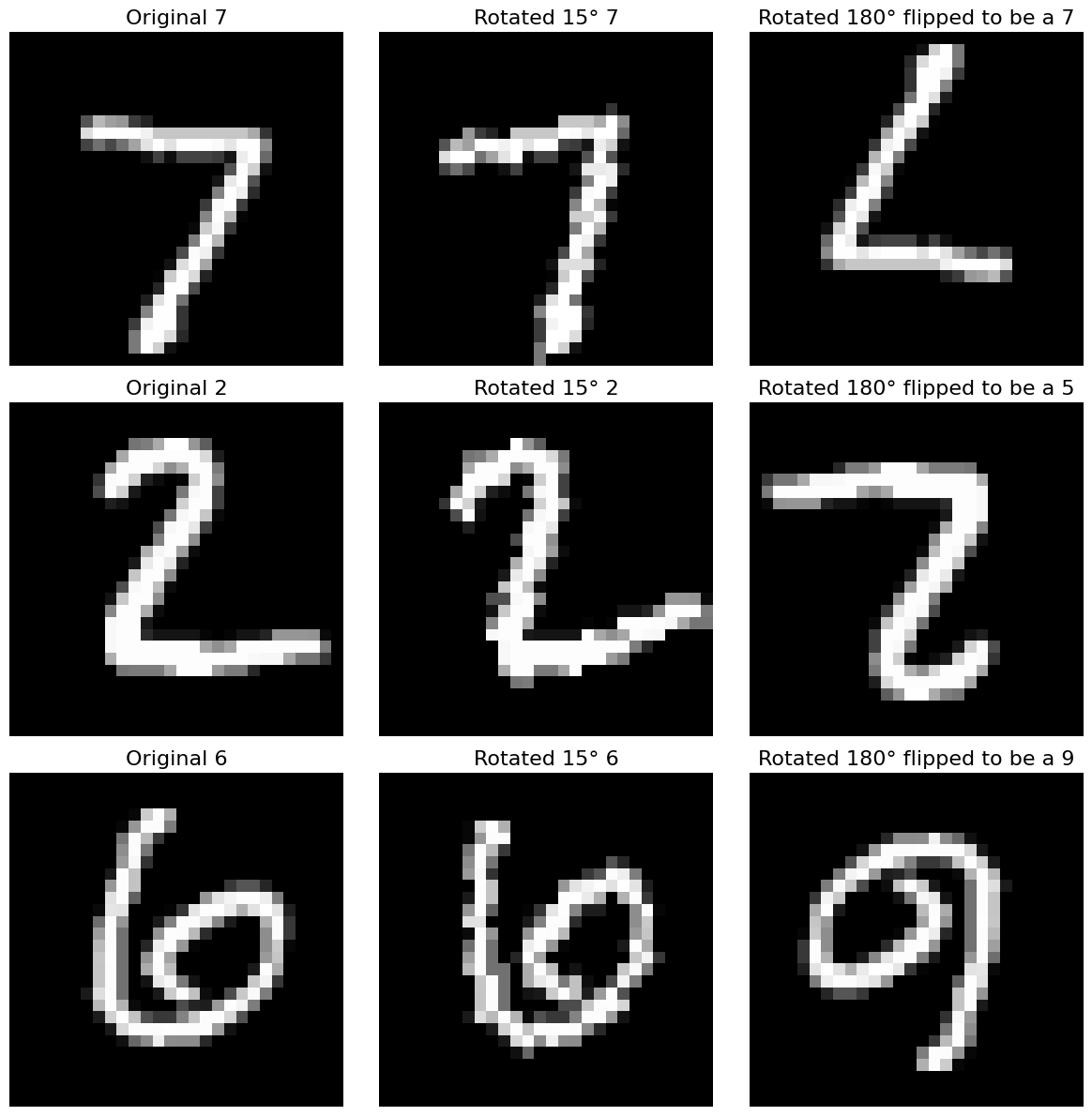}  
    \caption{Label preserving transformation vs non-label preserving}  
    \label{fig:non-label-preserve-examples}  
\end{figure}

We used the MNIST dataset for this experiment, applying a single hard transformation: a 180-degree rotation. This transformation does not merely augment the input but also changes the output label in a systematic way. For example, a '2' becomes a '5', and a '6' becomes a '9', with other digits not changing their labels. The details of this transformation are illustrated in \ref{fig:non-label-preserve-examples}.

\begin{itemize}
    \item \textbf{Transformation Details}:
    \begin{itemize}
        \item \textbf{Weak Augmentation}: A rotation of 15 degrees, which does not alter the label.
        \item \textbf{Strong Augmentation}: A rotation of 180 degrees, which changes the label (e.g., '2' to '5', '6' to '9').
        \item \textbf{Dataset}: The MNIST dataset, with only 8\% of the dataset used as the labeled set to simulate data scarcity.
    \end{itemize}
    \item \textbf{Algorithms Tested}: We evaluated multiple algorithms, including FixMatch, FullMatch, and MixMatch, both with and without adaptive retraining. The primary metrics assessed were accuracy and robustness on the non-label preserving set.
\end{itemize}

\begin{tcolorbox}[colback=white, colframe=black, title=Experimental Configuration]
\begin{itemize}
    \item \textbf{Dataset}: MNIST
    \item \textbf{Transformations}:
    \begin{itemize}
        \item \textbf{Weak Augmentation}: 15-degree rotation
        \item \textbf{Strong Augmentation}: 180-degree rotation (non-label preserving)
    \end{itemize}
    \item \textbf{Labeled Set}: 8\% of MNIST dataset
    \item \textbf{Algorithms Tested}: FixMatch, FullMatch, MixMatch
    \item \textbf{Configurations}: With and without adaptive retraining
    \item \textbf{Metrics}: Accuracy
    \item \textbf{Primary Focus}: Performance on non-label preserving set
\end{itemize}
\end{tcolorbox}

\begin{figure}[h]  
    \centering  
    \includegraphics[width=0.5\textwidth]{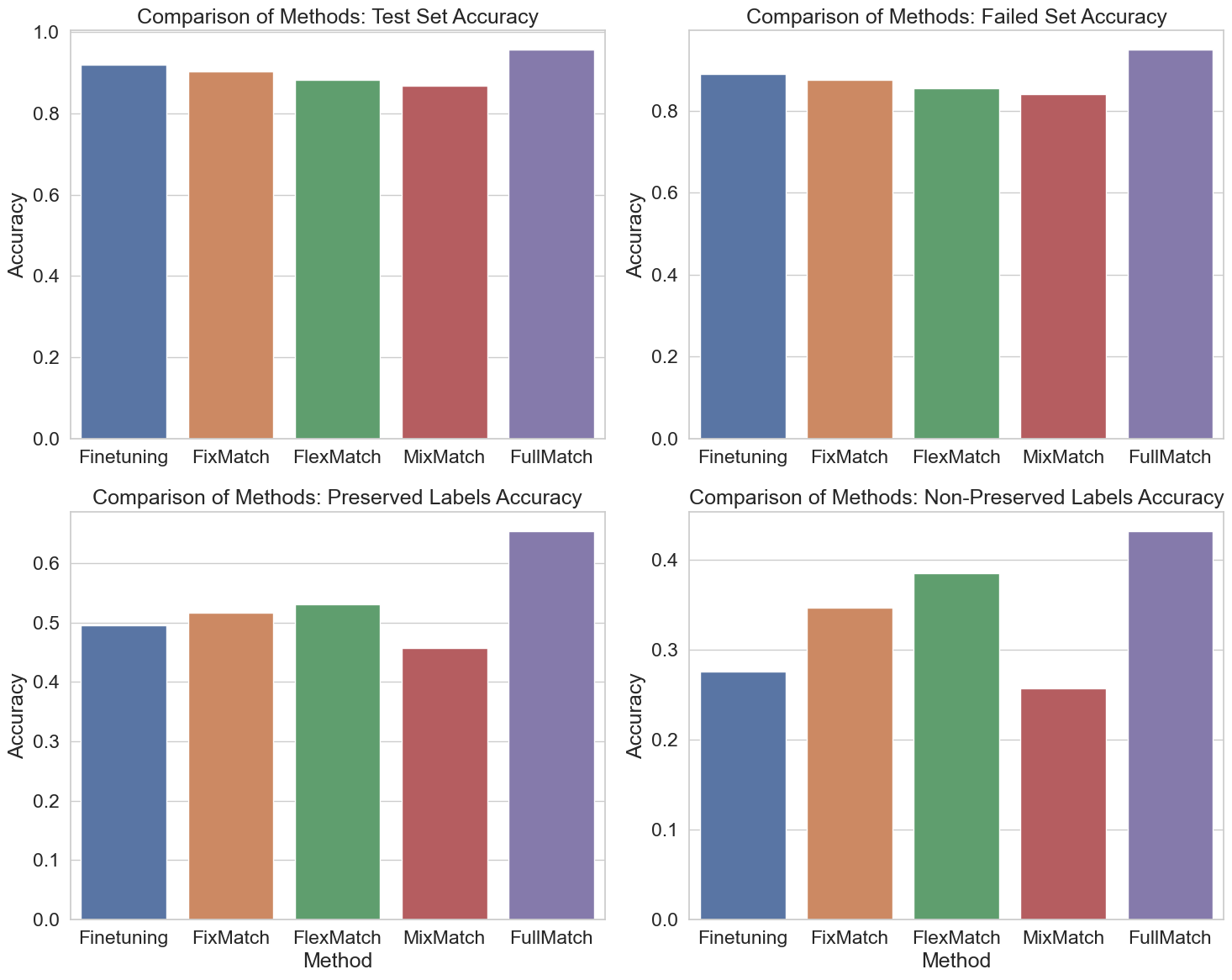}  
    \caption{Accuracy on different sets. Comparing the performance on label preserving and non-label preserving.}  
    \label{fig:non-label-preserving-accuracy}  
\end{figure}

\subsubsection{Results}
The results, shown in \ref{fig:non-label-preserving-accuracy}, highlight the performance differences among the algorithms on the non-label preserving test set, which consists of the test set fully augmented with a 180-degree rotation and adjusted labels.

\begin{itemize}
    \item \textbf{Performance on Non-Label Preserving Set}:
    \begin{itemize}
        \item \textbf{FullMatch}: This algorithm performed the best, showing superior robustness and accuracy on the non-label preserving set. FullMatch’s consistent performance across various experiments indicates its robustness.
        \item \textbf{FixMatch}: This algorithm also performed well but did not surpass FullMatch. It demonstrated solid robustness but was slightly behind FullMatch in terms of accuracy.
        \item \textbf{MixMatch}: This algorithm had the lowest performance by far. It struggled significantly with the non-label preserving transformations, indicating that it may not be well-suited for tasks requiring robust adaptation to such changes.
    \end{itemize}
\end{itemize}

In conclusion, this experiment demonstrates that FullMatch is the most robust algorithm when dealing with non-label preserving transformations. It outperforms other algorithms, including FixMatch and MixMatch, particularly under conditions of data scarcity and significant label alterations.

\subsection{Experiment 5: Adaptive Retraining vs. Static Metamorphic Retraining}

\subsubsection{Objective}
The objective of this experiment is to test the effectiveness of adaptive retraining compared to static metamorphic retraining. Specifically, we aim to determine whether adaptive retraining, which adjusts based on the outcomes of each retraining cycle, provides better accuracy and robustness compared to a static approach that does not differentiate between failed and successful tests.

\subsubsection{Experimental Setup}
This experiment was conducted using the CIFAR-10 dataset and a ResNet-50 model. To simulate a data-scarce environment, only 1\% of the dataset was used as labeled data. The experiment involved 30 robustness cycles, with two epochs per cycle, under two different retraining strategies:

\begin{tcolorbox}[colback=white, colframe=black, title=Experimental Configuration]
\begin{itemize}
    \item \textbf{Dataset}: CIFAR-10
    \item \textbf{Model}: ResNet-50
    \item \textbf{Labeled Data}: 1\% of CIFAR-10 dataset
    \item \textbf{Retraining Strategies}:
    \begin{itemize}
        \item \textbf{Adaptive Retraining}: Dynamic adjustment based on test outcomes
        \item \textbf{Static Metamorphic Retraining}: Uniform retraining strategy
    \end{itemize}
    \item \textbf{Robustness Cycles}: 30 cycles with 2 epochs per cycle
    \item \textbf{Metrics}: Accuracy, Robustness
    \item \textbf{Primary Focus}: Comparison between adaptive and static retraining strategies
\end{itemize}
\end{tcolorbox}

\begin{itemize}
    \item \textbf{Adaptive Retraining}: This method dynamically adjusts the retraining process based on the outcomes of previous cycles, making a distinction between failed and successful tests.
    \item \textbf{Static Metamorphic Retraining}: This method applies the same retraining strategy uniformly across all cycles without differentiating between the results of tests.
\end{itemize}

\begin{table}[h]
    \centering
    \begin{tabular}{|>{\centering\arraybackslash}m{2cm}|c|c|c|c|}
        \hline
        \rowcolor{white}
        & \multicolumn{2}{c|}{\textbf{Static Retraining}} & \multicolumn{2}{c|}{\textbf{Adaptive Retraining}} \\
        \cline{2-5}
        \rowcolor{white}
        & \textbf{Accuracy} & \textbf{Robustness} & \textbf{Accuracy} & \textbf{Robustness} \\
        \hline
        \textbf{FixMatch} & 50\% & 88\% & 55\% & 84\% \\
        \hline
        \textbf{FlexMatch} & 60\% & 79\% & 60\% & 83\% \\
        \hline
        \textbf{MixMatch} & 59\% & 74\% & 70\% & 85\% \\
        \hline
        \textbf{FullMatch} & 72\% & 62\% & 68\% & 87\% \\
        \hline
        \rowcolor[gray]{.9}
        \textbf{Average} & 60.5\% & 75.75\% & \cellcolor[HTML]{90EE90}63.25\% & \cellcolor[HTML]{90EE90}84.75\% \\
        \hline
    \end{tabular}
    \vspace{0.3cm}
    \caption{Comparison of Accuracy and Robustness between Static Retraining and Adaptive Retraining for \textbf{CIFAR-10}}
    \label{tab:static-retraining-vs-adaptive-retraining}
\end{table}

\subsubsection{Results}
The results, illustrated in \ref{tab:static-retraining-vs-adaptive-retraining}, highlight the differences in performance between adaptive retraining and static metamorphic retraining.

\begin{itemize}
    \item \textbf{Accuracy Improvement}:
    \begin{itemize}
        \item \textbf{Adaptive Retraining}: This method showed an improvement in accuracy over the Static Retraining algorithms. Its dynamic nature allows it to have greater accuracy.
    \end{itemize}

    \item \textbf{Robustness Comparison}:
    \begin{itemize}
        \item \textbf{Adaptive Retraining}: Has better robustness across the board.
        \item \textbf{Static Retraining}: Was more robust than the Base Method.
    \end{itemize}
\end{itemize}

In conclusion, this experiment validates the effectiveness of adaptive retraining over static metamorphic retraining. The adaptive approach’s ability to dynamically respond to test outcomes significantly improves accuracy and robustness. These findings emphasize the importance of flexibility and adaptability in training strategies, particularly in scenarios with limited labeled data and high demands for model robustness.

\section{Individual Contributions}

The concept and framework for the Metamorphic Retrainer were collaboratively developed. Once established, we efficiently divided the work to ensure all parts of the architecture were completed asynchronously and on time.

\begin{itemize}
\item \textbf{Youssef Sameh Mostafa:} Authored and wrote the Metamorphic retrainer, developed an extensible integration of models, datasets, MRs and integrated the non-label preserving transformations with the semi-supervised algorithms.
\item \textbf{Karim Lotfy:} Integrated the project with GeMTests, implemented the label-transformation extraction, and carried out the Robustness Metamorphic tests. Additionally, he wrote the Metamorphic Tester.
\item \textbf{Aziz Said Togru:} Created the training script to execute the different semi-supervised algorithms and the normal training procedure. Authored and validated the FixMatch, FlexMatch, MixMatch, and FullMatch algorithms.
\end{itemize}

Throughout the project, we collaborated closely, meeting twice a week—once with Simon and once (sometimes multiple times) among ourselves. These meetings were dedicated to discussing and brainstorming the architecture and experimental approaches, ensuring a cohesive and well-thought-out execution of our project.

\section{Conclusion}
In this study, we explored various retraining methods for semi-supervised algorithms, focusing on how different augmentation strategies impact model performance. By setting a fixed number of iterations or employing a stopping criterion based on performance thresholds, we aimed to optimize the efficiency and effectiveness of the training process. Our experiments compared three primary retraining approaches: using base data augmentations, adaptive methods with strong augmentations derived from failed tests, and combining weak and strong augmentations based on metamorphic relations.

The results demonstrated that each method has distinct advantages, with adaptive augmentation strategies showing particular promise in enhancing model robustness and accuracy. By specifying failed tests as strong augmentations, the adaptive method effectively addressed specific weaknesses in the model, leading to significant performance improvements.

In conclusion, our study highlights the importance of carefully designed retraining methods and augmentation strategies in semi-supervised learning. By continuing to innovate and refine these approaches, we can enhance the capability of ML models to perform effectively in real-world scenarios with limited labeled data.

\section{Future Work}

Future work will involve further refining these retraining methods and exploring their applicability to a broader range of models and tasks. One potential direction is to develop more sophisticated criteria for stopping iterations, potentially incorporating dynamic performance metrics that adjust based on real-time feedback during training. Additionally, investigating the use of automated machine learning (AutoML) techniques to optimize augmentation strategies could provide a more streamlined and scalable approach to model retraining.

Another promising avenue for future research is the integration of these retraining methods with other advanced techniques, such as transfer learning and domain adaptation. By combining the strengths of semi-supervised learning with these techniques, we can potentially achieve even greater improvements in model performance and generalization across diverse datasets and domains.

\section{Acknowledgment}

This project was completed as part of the master's practical course "Advanced Testing of Deep Learning Models: Towards Robust AI" at the Chair of Software and Systems Engineering at the Technical University of Munich, led by Prof. Dr. Alexander Pretschner, Vivek V. Vekariya, and Simon Speth. We would like to thank the organizers for giving us the opportunity to work on projects like this and to make a contribution.

Additionally, we would like to give special thanks to our supervisor, Simon Speth, who personally guided us through this project. His weekly support in refining our ideas and execution was invaluable in steering us in the right direction.

\bibliographystyle{plain}
\bibliography{references}

\end{document}